\documentclass[twoside,leqno,twocolumn]{article}
\usepackage{kbordermatrix}
\usepackage{hyperref}
\usepackage{cite}
\usepackage{algorithmic}
\usepackage{amsmath}
\usepackage{amsfonts}
\usepackage{caption}
\usepackage{dsfont}
\usepackage{float}
\usepackage{graphicx}
\graphicspath{ {} }
\DeclareGraphicsExtensions{.eps}
\usepackage[titlenumbered,ruled]{algorithm2e}
\usepackage{mathtools}

\DeclarePairedDelimiter\floor{\lfloor}{\rfloor}
\usepackage{ltexpprt}

\begin{document}
\def\bx{\mbox{\boldmath $x$}}
\def\be{\mbox{\boldmath $e$}}
\def\bu{\mbox{\boldmath $u$}}
\def\bv{\mbox{\boldmath $v$}}
\def\bw{\mbox{\boldmath $w$}}
\def\bQ{\mbox{\boldmath $K$}}
\def\bA{\mbox{\boldmath $A$}}
\def\bE{\mbox{\boldmath $E$}}
\def\xpi{{\bx}_{i}^{+}}
\def\xnj{{\bx}^{-}_{j}}
\def\scd{\cal D}
\def\half{\frac{1}{2}}
\def\bQb{\beta^T {\bQ} \beta}
\def\bjQbj{\beta^T_J {\bQ}_{J,J} \beta_{J}}
\def\Qbi{{({\bQ} \beta)}_{i}}
\def\QbiJ{{({\bQ} \beta)}_{i,J}}
\def\Qbj{{({\bQ} \beta)}_{j}}
\def\QbjJ{{({\bQ} \beta)}_{j,J}}
\def\QlJ{{\bQ}_{\cdot,J}}
\def\QbJ{{\bQ}_{J,J} \beta_{J}}

\title{\Large A Sparse Non-linear Classifier Design Using AUC Optimization\thanks{}}
\author{Vishal Kakkar\thanks{Computer Science \& Automation, IISc Bangalore, India.} \\
\and
Shirish K. Shevade\thanks{Computer Science \& Automation, IISc Bangalore, India.}
\and
S Sundararajan\thanks{Microsoft Research, Bangalore, India.}
\and
Dinesh Garg \thanks{IIT Gandhinagar, India. }
}
\date{}

\maketitle



\begin{abstract} \small\baselineskip=9pt 
AUC (Area under the ROC curve) is an important performance measure for applications where the data is highly imbalanced. Learning to maximize AUC performance is thus an important research problem. Using a max-margin based surrogate loss function, AUC optimization problem can be approximated as a pairwise rankSVM learning problem. Batch learning methods for solving the kernelized version of this problem suffer from scalability and may not result in sparse classifiers. Recent years have witnessed an increased interest in the development of online or single-pass online learning algorithms that design a classifier by maximizing the AUC performance. The AUC performance of nonlinear classifiers, designed using online methods, is not comparable with that of nonlinear classifiers designed using batch learning algorithms on many real-world datasets. Motivated by these observations, we design a scalable algorithm for maximizing AUC performance by greedily adding the required number of basis functions into the classifier model. The resulting sparse classifiers perform faster inference. Our experimental results show that the level of sparsity achievable can be order of magnitude smaller than the Kernel RankSVM model without affecting the AUC performance much.\end{abstract}

\section{Introduction.}
In binary classification, a classifier is often trained by optimizing a performance measure such as accuracy. If the data is highly imbalanced, accuracy may not be a good measure to optimize. The all-positive or all-negative classifier may achieve good classification accuracy. But, this will result in misclassification of some important or rare events which typically belong to a minority class. Situations for which datasets are imbalanced are not uncommon in real-world applications and in such cases, classifiers are designed by optimizing measures other than accuracy \cite{fawcett2001using}.

Support Vector Machines (SVMs) have been very effective on several real-world problems. Standard SVM formulations for binary classification problem assumes that misclassification costs are equal for both the classes. Therefore, SVMs are not suitable if the data is strongly imbalanced. Lin et. al.~\cite{lin2002support} proposed a simple extension of SVMs by using different penalization of positive and negative examples. This approach is useful if misclassification costs are known, which is typically not the case in practice. It is thus necessary to use a different measure for learning from imbalanced data.

AUC (Area Under ROC Curve) \cite{metz1978basic}, \cite{hanley1983method} is an important performance measure and its optimization has been very effective, especially when class distributions are heavily skewed. However, computing the AUC is a costly operation as the AUC is written as a sum of pairwise losses between examples from different classes, which is quadratic in the number of training set examples. Further, the AUC is not a continuous function on the training set. This makes the optimization of AUC a challenging task.

Many algorithms have been designed to optimize AUC using surrogate loss functions (Herschtal et. al.~\cite{herschtal2004optimising}, Joachims et. al.~\cite{joachims2005support}, Rudin et. al.~\cite{rudin2009margin}, Kotlowski et. al.~\cite{kotlowski2011bipartite}, Zhao et. al.~\cite{zhao2011online}). Due to the high computational demands of the AUC or its variants, most of these algorithms are either one-pass algorithms or online algorithms which rely on sampling. Zhao et. al.~\cite{zhao2011online} proposed an online AUC algorithm (OAM) which is based on the idea of reservoir sampling. This idea helps to represent all the received examples by the examples stored in buffers of fixed size. Gao et. al.~\cite{gao2013one} proposed a regression based algorithm for one-pass AUC (OPAUC) optimization. This algorithm maintains only the first and second order statistics of training data in memory, thereby resulting in a storage requirement which is independent of the training dataset size. Both these algorithms learn linear classifiers and are not directly suitable to design complex nonlinear decision boundaries, typically possible by using kernel classifiers. 

Calders et. al.~\cite{calders2007efficient} proposed the use of polynomial approximations for the AUC, which can be computed in only one scan over the dataset. This approximation was used to design a linear classifier. Yang et. al.~\cite{yang2013online}  proposed an online learning algorithm to optimize the AUC score by learning a nonlinear decision function via the kernel trick. This method, called online imbalanced learning with kernels (OILK), maintains a buffer to store the informative support vectors. Two buffer update policies, first-in-first-out and reservoir sampling were investigated. As the cost of determining the AUC score is very large, most of these algorithms avoid the exact computation of the AUC score and resort to online or one-pass approaches by making use of buffers to store the relevant information. Although the storage requirements are reduced for such methods, generalization performance of the resulting classifiers is not comparable with that of the nonlinear classifiers designed using batch learning algorithms on many real world datasets.

More relevant to the work in this paper is the large-scale Kernel RankSVM algorithm proposed by Kuo et. al.~\cite{kuo2014large}. This algorithm, though designed for solving a ranking problem, can be extended to solve the AUC optimization problem. However, kernel evaluations are a bottleneck in training Kernel RankSVM. To alleviate this problem, it was proposed to store the full kernel matrix. Although this reduces repeated kernel evaluations, storage of the full kernel matrix is an issue if the dataset sizes are very large, as it requires $O(l^2)$ storage. Further, Kernel RankSVM may result in a model which uses a large number of support vectors, thereby incurring high inference cost.

Motivated by the above observations, we propose an algorithm to learn sparse models
for maximizing AUC using a max-margin based surrogate loss function. Our experimental results show that the level of sparsity achievable can be order of magnitude smaller than the Kernel RankSVM model without affecting the AUC performance much. This helps to achieve significant speed-up during prediction. Due to the nature of our algorithm, parallelization is possible and we demonstrate that significant training speed-up is achievable by using a multi-core version of the algorithm.

Notation: Here we discuss the notations we have used in our work. All vectors will be column vector and row vectors will be denoted by a superscript, $^{T}$. 2-norm of the vector x is denoted by $\|x\|$. $|J|$ denotes the cardinality of the set J. $\bQ$ denotes the kernel matrix. $\QlJ$ refers to the submatrix of $\bQ$ made of all the $l$ rows and the columns indexed by J. $\bQ_{i,J}$ refers to the ith row of the matrix \bQ. $\bQ_{JJ}$ refers to the submatrix of $\bQ$ made of the rows indexed by J and the columns indexed by J.

\section{Problem Definition}
Let the training data be denoted by $\scd$ $=P \cup N$, where $P = \{\xpi, +1\}^p_{i=1}$, $N = \{\xnj, -1\}^n_{j=1} $ and ${\bx}^+_i, {\bx}^{-}_j \in R^d$. We will denote $q^{th}$ training set example as ${\bx}_q$. Let $T$ denote the index set of pairs of positive and negative instances in $\;\scd$. Clearly, $\;|T| = pn$. Let $\; l = p + n$. Without loss of generality, we assume that $\;p \ll n$. 

We assume that the non-linear decision function $\;f(\cdot)$ is an element of a Reproducing Kernel Hilbert Space (RKHS). That is, $\;f$ is a linear combination of kernel functions,
\begin{equation}
\label{eq:6}
f({\bx}) = {\bw} \cdot \phi({\bx}) = \sum^l_{q=1} \beta_q \phi({\bx}_q)
\cdot \phi({\bx}) = \sum^l_{q=1} \beta_q k({\bx}, {\bx}_q),
\end{equation}
where $\phi(\cdot)$ maps the data into high dimensions and $k(\cdot,\cdot)$ denotes a kernel function. The AUC score of the function $\;f$ on the dataset $\scd$ is defined as
\small
\begin{eqnarray}
\label{eq:7}
\mbox{AUC}(f) & = & \frac{\sum^p_{i=1} \sum^n_{j=1} \; I(f(\xpi) >
f(\xnj))}{pn} \notag \\
& = & 1 - \frac{\sum^p_{i=1} \sum^n_{j=1} \; I(f(\xpi) \leq f(\xnj))}{pn}
\end{eqnarray}
\normalsize
where $I(\cdot)$ is the indicator function which outputs $\;1$ if the argument is true and $\;0$ otherwise. Thus maximizing AUC($f$) is equivalent to minimizing ${\sum^p_{i=1} \sum^n_{j=1} \; I(f(\xpi) \leq f(\xnj))}$. Writing $f({\bx}_i) = ({\bQ} \beta)_i$ and using a max-margin based surrogate loss function (a hinge or a squared hinge loss), we get the following two regularized formulations corresponding to the two loss functions:
\small
\begin{equation}
\label{eq:8}
\min_{\beta \in R^l} \;\; \half \bQb + C \sum_{(i,j) \in T} \;\max(0, 1- {\Qbi} + {\Qbj})
\end{equation}
and
\begin{equation}
\label{eq:9}
\min_{\beta \in R^l} \;\; \half \bQb + \frac{C}{2} \sum_{(i,j) \in T}\;
\max(0, 1 - {\Qbi} + {\Qbj})^2
\end{equation}
\normalsize
where $C$ is a positive hyperparameter that controls the error.

In this work, we focus on problem (\ref{eq:9}) as it is a continuously differentiable function and devise an efficient algorithm to solve it. Unlike typical classification problems where the loss function can be calculated for every single training set example, the second term in (\ref{eq:9}) involves losses defined over pairs of examples from different classes. This makes the problem (\ref{eq:9}) more challenging.

\section{Related Work}
We now briefly review some of the related works for AUC optimization.

Many online algorithms have been proposed to learn a linear classifier by maximizing the AUC score. These algorithms include Online AUC Maximization (OAM) ~\cite{zhao2011online} and Adaptive Online AUC Maximization (AdaOAM) ~\cite{ding2016adaptive}. AUC optimization in online learning is a challenging task as the computation of the AUC score involves the sum of pairwise losses between instances from opposite classes. To tackle this challenge, online learning uses the idea of buffer sampling ~\cite{zhao2011online}~\cite{kar2013generalization}. A fixed size buffer is used to represent all the observed data by storing some randomly sampled examples in it. Kar et. al.~\cite{kar2013generalization} introduced the idea of stream subsampling with replacement as the buffer update strategy. Although these online algorithms have demonstrated good AUC performance by using simple online gradient descent approaches, they do not use the geometrical knowledge of the observed data. AdaOAM overcomes this limitation by employing an adaptive gradient method that exploits the knowledge of historical gradients. Its variant, SAdaOAM was proposed to design a sparse model in online AUC maximization task. Gao et. al.~\cite{gao2013one} proposed a one-pass optimization algorithm by considering square loss for the AUC optimization task. Due to the use of squared error loss, the algorithm only needs to store the first and second order statistics for the observed data.

A main drawback of the online methods discussed above is that they learn a linear classifier and do not exploit the learning power of kernel methods. To  address this issue, yang et. al.~\cite{yang2013online} investigated Online Imbalanced Learning with Kernels (OILK) where informative support vectors are stored in the buffer. Two buffer update strategies, First-In-First-Out (FIFO) and Reservoir Sampling (RS) were investigated. By conducting experiments on real-world datasets, it was demonstrated that the kernel methods for AUC maximization performed better than their linear classifier counterparts. The proposed method~\cite{yang2013online} is however an online algorithm.

Joachims~\cite{joachims2005support} presented a structural SVM framework for optimizing AUC in a batch mode. By formulating (\ref{eq:9}) as a 1-slack structural SVM problem, Joachims~\cite{joachims2005support} solved its dual problem by a cutting plane method. The method, though initially designed for linear classifiers, can be easily extended to nonlinear classifiers. Numerical experiments showed that, for ranking learning problems, this method is slower than others state-of-the-art methods that solve (\ref{eq:9}) directly~\cite{kuo2014large}~\cite{lee2014large} .

Learning to rank is an important supervised learning problem and has application in a variety of domains such as information retrieval and online advertising. Treating the all instances query number same, the set of preference pairs will be same as T and the Kernel rankSVM  problem, discussed in~\cite{kuo2014large} is same as (\ref{eq:9}). Kuo et. al.~\cite{kuo2014large} used trust region Newton method to solve this problem. This method stores the full kernel matrix as repeated kernel evaluations are bottleneck in Kernel rankSVM. This method has two drawbacks: 1) it is not scalable as the memory requirement is prohibitively high for large datasets, and 2) the learned model is not sparse resulting in computationally expensive predictions.

\def\bQ{\mbox{\boldmath $K$}}

\section{Our Approach: Sparse Kernel AUC}
Our aim is to learn a sparse nonlinear classifier model for a binary classification problem with imbalanced data distributions for the two classes. 
We now discuss our approach to solve (\ref{eq:9}). A similar problem formulation was used in \cite{kuo2014large} to solve the problem of learning to rank and the algorithm designed there is also applicable to our setting. Kuo et. al.~\cite{kuo2014large}  alleviated the difficulty of computing the loss term, which involves summation over preference pairs, by using order-statistic trees. Although the cost of computing the required quantities was reduced to $O(l\log l)$ from $O(l^2)$, the kernel evaluations amount to \textbf{$O(ld)$} time, which can be reduced to $(O(l)$ if the kernel matrix ${\bQ}$ is maintained throughout the optimization algorithm. In their implementation, Kuo et. al.~\cite{kuo2014large} store the full kernel matrix ${\bQ}$ which is a dense matrix of size $l \times l$. However, for large datasets it is impractical to store the full kernel matrix ${\bQ}$ in the main memory. Further, for such huge datasets, the resulting classifier may not be sparse, thereby making the inference slow. It is therefore desired to devise a different approach to solve (\ref{eq:9}) and design a sparse classifier.

Motivated by the success of the matching pursuit approach, presented by Keerthi et. al.~\cite{keerthi2006building}, to design sparse SVM classifiers, we propose a new and efficient algorithm to solve (\ref{eq:9}) using matching pursuit ideas. The algorithm requires to compute and maintain the kernel matrix of size $\; l \times d_{max}$ (where $d_{max}$ is the user specified positive parameter whose value can be about $5-10\%$ of the dataset size $l$) which helps to reduce the memory requirement considerably. For the dataset with $l = 49,990$, we observed that $d_{max} \approx 200$ was sufficient to achieve very good AUC performance on the test set. 

We also demonstrate that efficient computations of the objective function in (\ref{eq:9}), gradient and Hessian-vector product computations are done by using simple techniques like sorting, binary search and hashing \cite{knuthart} and do not require the use of sophisticated data structures such as order-statistic trees. As our experimental results show, the proposed approach is faster than the approach of Kuo et. al.~\cite{kuo2014large} applied to the AUC maximization problem and achieves comparable generalization performance using small number of support vectors.

We now discuss the key components of our proposed algorithm.

\subsection{Reformulation}
Borrowing the ideas presented in \cite{keerthi2006building}, we maintain a set of greedily chosen kernel basis functions to design a sparse non-linear
classifier. The cardinality of this set is denoted by $d_{max}$, a user specified positive integer. Let $\;J$ denote the index set of these basis functions. In our experiments, we choose $\;J \subset \{1,2,\ldots,l\}$. Having defined the set $J$, the parameter vector ${\bw}$ in (\ref{eq:6}) can be represented as
\[
{\bw} = \sum_{q \in J} \; \beta_q \phi({\bx}_q)
\]
and the problem formulation in (4) can be written as
\footnotesize
\begin{equation}
\label{eq:10}
\min_{\beta_J \in R^{|J|}} \bE(\beta_{J}) \equiv \half \bjQbj + \frac{C}{2} \sum_{\mathclap{(i,j) \in T}} \max(0, 1 - \bQ_{i,J}\beta_{J} + \bQ_{j,J}\beta_{J})^2
\end{equation}
\normalsize
Note that the Kernel rankSVM algorithm solves the following problem;
\begin{equation}
\label{eq:B}
\min_{\beta \in R^{l} } \frac{1}{2} \beta^{T}\bQ\beta + \frac{C}{2}\sum_{\mathclap{(i,j) \in S}} max(0,1-(\bQ\beta)_{i} + (\bQ\beta)_{j})^{2}
\end{equation} 
where $S = \{(i,j)|q_{i}=q_{j},y_{i}>y_{j}\}$ is the set of preference pairs for queries $q$. This problem requires either to store the full kernel matrix $\bQ$ or requires many kernel evaluations, which become a bottleneck. On the other hand, the solution to our problem (\ref{eq:10}) requires to store the matrix of size $l \times d_{max}$, which makes it scalable.

In this work, we solve (\ref{eq:10}) using matching pursuit ideas~\cite{vincent2002kernel},~\cite{keerthi2006building}. In this approach, starting with $J = \phi$, a training set example is chosen from the set $\;\{1,2,\ldots,l) \setminus J$ such that its inclusion in the set $J$ results in a maximum improvement in the objective function. The optimization problem is then solved with respect to $\beta_J$. This procedure is repeated till $\;|J| = d_{max}$ holds true. Algorithm 1 gives the pseudo-code of this procedure. Step 5 of this algorithm is computationally expensive and  in section V, we will discuss some approaches to make it efficient\\
\normalsize

The efficiency of this algorithm depends on the efficient computation of the objective function value $\bE(\beta_{J})$, its gradient $\nabla \bE(\beta_{J}) $ and Hessian-vector product $\nabla^{2} \bE(\beta_{J})\bv$ for any vector $R^{|J|}$. If $\bA$ a pairwise indexing matrix and $\bA_{\beta_{J}}$ denotes the indexing matrix of violating pairs, which contribute to the loss function, then by defining 
\begin{equation}
\label{eq:c}
\bu_{\beta_{J}} = \bA_{\beta_{J}}^{T}\bA_{\beta_{J}}\QlJ \beta_{J},
\end{equation}
the problem in (\ref{eq:10}) can be re-written as
\footnotesize
\begin{align}
\label{eq:A}
\min_{\beta_{J}} \; \bE(\beta_{J}) \equiv  \half \bjQbj + \frac{C}{2} ( \beta_{J}^{T}\QlJ^{T}( \bu_{ \beta_{J}} - 2\bA_{\beta_{J}}^{T}\be_{\beta_{J}} )+p_{\beta_{J}}).
 \end{align}
\normalsize
where $p_{\beta_{J}}$ is the number of violating pairs (details given in appendix).
This rewriting helps in computing $\bE(\beta_{J})$, $\nabla \bE(\beta_{J}) $ and  $\nabla^{2} \bE(\beta_{J})\bv$ efficiently as all of these quantities require the computation of $\bu_{\beta_{J}}$. By defining
\small
\begin{equation}
\label{eq:C}
SV(\beta_{J}) =  \{(i,j) \in T \; | \;1 - {\bQ_{i,J}\beta_{J}} + {\bQ_{j,J}\beta_{J}} >0 \}
\end{equation}
\normalsize
and 
\begin{center}
$ SV_{i}^{+}(\beta_{J}) \equiv  \{ j \mid (j,i) \in SV(\beta_{J})  \} $, $l_{i}^{+}(\beta_{J}) \equiv |SV_{i}^{+}(\beta_{J})|$, $\gamma_{i}^{+}(\beta_{J},\bv) \equiv \sum_{j \in SV_{i}^{+}(\beta_{J})} \bQ_{j,J}^{T}\bv$,
$ SV_{i}^{-}(\beta_{J}) \equiv  \{ j \mid (i,j) \in SV(\beta_{J})  \} $, $l_{i}^{-}(\beta_{J}) \equiv |SV_{i}^{-}(\beta_{J})|$ , $\gamma_{i}^{-}(\beta_{J},\bv) \equiv \sum_{j \in SV_{i}^{-}(\beta_{J})} \bQ_{j,J}^{T}\bv$.\\
\end{center}
\normalsize
one compute $\bu_{\beta_{J}}$ efficiently, as can be seen from equation (\ref{eq:17}) in appendix.

\renewcommand{\algorithmicrequire}{\textbf{Input:}}
\renewcommand{\algorithmicensure}{\textbf{Output:}}

\begin{algorithm}
\SetAlgoLined
\label{alg:2}
\begin{algorithmic}[1]
\REQUIRE $\scd$ $ = \{\xpi, +1\}^p_{i=1} \cup \{\xnj, -1\}^n_{j=1} $, C, $d\_max$
\ENSURE $J$, $\beta_{J}$
\STATE $J = \phi$
\WHILE{$|J| < d_{max}$}
\STATE select a new basis function $j^{*}$ which gives a maximum decrease in the objective function $\bE_{\beta_{J}}$
\STATE $J = J \cup \{ j^{*} \}$
\STATE Optimize the objective function w.r.t $\beta_{J}$
\ENDWHILE
\end{algorithmic}
\caption{{Sparse Classifier Design Algorithm}}
\end{algorithm}

Lee et. al.~\cite{lee2014large} and Airola et. al.~\cite{airola2011training} used order-statistic trees to efficiently compute the $l_{i}^{+}(\beta_{J})$ and  $l_{i}^{-}(\beta_{J})$ for rankSVM. The problem of maximizing AUC does not require order-statistic trees. It is enough to use sorting, searching and hashing methods. The details are given in Algorithm 2.\\

For a given $\beta_{J}$, we define the set of ordered pairs which contributes to the empirical loss of objective function in (\ref{eq:10}) as $SV(\beta_{J})$. For every example in the training set, by finding out the set of violating examples of the other class ($SV^{+}$ and $SV^{-}$) and the sum of $\QlJ$ for those examples ($\gamma^{+}$ and $\gamma^{-}$), we can compute the empirical loss term in (\ref{eq:10}). These computations can be done efficiently by using sorting (Steps 1-7), hashing (Steps 9-12) and searching (Steps 15-26 ). The complexity of this algorithm is $O(l(d_{max}+ \log l))$\footnote{$ld_{max}$ is for computing $\QlJ$ and $l \log l$ is for sorting}, which is better than naive computation of pairwise losses in (\ref{eq:10}). Further, in our experiments we implemented steps 15-26 of Algorithm 2 in multi-core setting. Empirical evaluation, discussed in the next section, shows that this resulted in a significant speed up of our algorithm.

\begin{algorithm}[th!]
\begin{algorithmic}[1]
\SetAlgoLined
\label{alg:1}
\REQUIRE $\QlJ$, $\beta_{J}$, $\bv$, P, N
\ENSURE  $l_{i}^{+}(\beta_{J})$, $l_{i}^{-}(\beta_{J})$, $\gamma_{i}^{+}(\beta_{J},\bv)$ and  $\gamma_{i}^{-}(\beta_{J},\bv)$
\STATE scoreP = $zeros(2,|P|)$, scoreN = $zeros(2,|N|)$
\STATE scoreP[1] = $\bQ_{iJ} * \beta_{J}$, for all $i \in P$
\STATE scoreP[2] = $\bQ_{iJ} * \bv$, for all $i \in P$
\STATE sort scoreP w.r.t to first row
\STATE scoreN[1] = $\bQ_{jJ} * \beta_{J}$, for all $j \in N$
\STATE scoreN[2] = $\bQ_{jJ} * \bv$, for all $j \in N$
\STATE sort scoreN w.r.t to first row
\STATE scorePsum = scoreP[2], scoreNsum = scoreN[2] 
\FOR{$i=2$ \TO $|P|$}
\STATE scorePsum[i] = scorePsum[i] + scorePsum[i-1]
\ENDFOR
\FOR{$i=|N|-1$ \TO $1$ }
\STATE scoreNsum[i] = scoreNsum[i] + scoreNsum[i+1] 
\ENDFOR
\FOR{$i=1$ \TO $|P|$} 
\STATE score = $(K_{i,J} * \beta_{J})$ -1
\STATE find the index k of scoreN using binary search s.t. $ scoreN[k-1] < score \leq scoreN[k]$
\STATE $l_{i}^{-}(\beta_{J}) = length(k:|N|)$ 
\STATE $\gamma_{i}^{-}(\beta_{J},\bv) = scoreNsum[k]$
\ENDFOR
\FOR{$j=1$ \TO $|N|$}
\STATE score = $(K_{j,J} * \beta_{J})$ +1
\STATE find the index k of scoreP using binary search s.t. $ scoreP[k] \leq score < scoreP[k+1]$
\STATE $l_{j}^{+}(\beta_{J}) = k$
\STATE $\gamma_{j}^{+}(\beta_{J},\bv) = scorePsum[k]$
\ENDFOR
\caption{Calculating $l_{i}^{+}(\beta_{J})$, $l_{i}^{-}(\beta_{J})$, $\gamma_{i}^{+}(\beta_{J},\bv)$, and  $\gamma_{i}^{-}(\beta_{J},\bv)$}
\end{algorithmic}
\end{algorithm}

\subsection{Basis Selection:}
We now discuss how to choose the kernel basis functions for a given problem. Our approach is greedy~\cite{keerthi2006building} and starts with an empty set $J$. A training set example is chosen from the set $\;\{1,2,\ldots,l) \setminus J$ such that its inclusion in the set $J$ results in a maximum improvement in the objective function. The optimization problem is then solved with respect to $\beta_J$. This procedure is repeated till $\;|J| = d_{max}$ holds true. Algorithm 1 gives the details of this procedure. The efficiency of this procedure depends on the optimization method used to solve (5). We discuss the two methods to add basis functions in the set J.\\

\subsubsection{Method 1}
A straightforward method is to choose every $q \in \scd \setminus$ $J$ and include in J (i.e., $J = J \cup q$), optimize (\ref{eq:10}) fully using $(\beta_{J},\beta_{q})$ and calculate the improvement in the objective function. Let it be $\bE_{q}^{~}$. Choose \[j^{*} = arg\min_{q \in D \setminus J} \bE_{q}\] in Step 3 of Algorithm (1). But solving (\ref{eq:10}) fully, $|\scd$ $\setminus J| \approx O(l)$ times is computationally expensive. 
Instead of choosing every $q \in$ $|\scd \setminus$ $J|$, we can work with a smaller subset of size $\kappa$ in $\scd$ $\setminus J$. Smola in \cite{smola2001sparse} suggested this random subset choice (of size 59), for Gaussian process regression successfully. But even with $\kappa$ number of random examples, this method of selection is still quite computationally heavy.\\

\subsubsection{Method 2}
In method 1, we solve a $|J|+1$ dimensional problem, optimizing $(\beta_{J},\beta_{q})$ to solve (\ref{eq:10}) completely. Instead it may be good idea to fix $\beta_{J}$ and solve (\ref{eq:10}) for $\beta_{q}$, to determine $\bE_{q}$. This problem is easy to solve as it is a one-dimensional problem.
\small
\begin{align}
\label{eq:25}
&\min_{\beta{q}} \frac{1}{2} 
  \left(
  \begin{array}{cc} 
  \beta_{J}^{T} & \beta_{q}  
  \end{array} 
  \right)  
  \left( 
  \begin{array}{cc}
\bQ_{J,J} & \bQ_{J,q}  \\
\bQ_{q,J} & \bQ_{q,q}  
\end{array} 
\right)
 \left(
 \begin{array}{c}
\beta_{J}   \\
\beta_{q}  
\end{array} 
\right) 
\text{ }+ \notag \\ 
&\frac{C}{2} \sum_{\mathclap{(i,j) \in T }} max (0,1-(\bQ_{i,J}-\bQ_{i,J})\beta_{J} - (\bQ_{i,q}-\bQ_{j,q})\beta_{q})^2 
\end{align} 
\normalsize
Keerthi et. al.~\cite{keerthi2006building} showed to solve this one dimensional problem in $O(l)$ time for SVM by using Newton-Raphson type iterations. So in our case complexity of solving this one dimensional problem will be $O(l(d_{max}+ \log l))$. Here also instead of choosing $q \in$ $\scd$ $\setminus J$, we choose q in random sample of size $\kappa$. 

\subsection{Truncated Newton Optimization Method}
The function $\bE(\beta_{J})$ (\ref{eq:10}) which we consider, can be optimized using second order optimization method. We use Truncated Newton Optimization Method (TRON) instead of Classical Newton Method, because in the classical newton method the update step is $\beta = \beta - H^{-1}g$. Computation of the Hessian and its inverse is a computational intensive task. Therefore, to reduce the computation time, we uses Truncated Newton Iteration to optimize current $\beta_{J}$. We use the linear conjugate gradient iteration in this method to approximate the $H^{-1}g$ which uses the Hessian-vector product for some vector \bv. As we discussed in the Section IV that, we can compute Hessian-vector product efficiently with overall complexity $$O(l(d_{max}+ \log l))$$.
\begin{algorithm}[h!]
\SetAlgoLined
\label{alg:4}
\begin{algorithmic}[1]
\REQUIRE J, C, current $\beta_{J}$
\ENSURE Optimized $\beta_{J}$
\STATE $\beta_{J}^{0} = \beta_{J}$, k = 0
\WHILE {$\beta_{J}^{k}$ is not optimal for obj fun}
\STATE Get direction d using linear CG (which requires Hessian-vector product $\&$ gradient) method
\STATE Find step size t using line search
\STATE Update $\beta_{J}^{k+1}  = \beta_{J}^{k} + td $
\STATE Set k=k+1
\ENDWHILE
\end{algorithmic}
\caption{Truncated Newton Iteration}
\end{algorithm}

We have not discuss the details of the linear conjugate gradient iteration. The details of the steps involved in linear conjugate gradient algorithm can be found in \cite{barrett1994templates}. There are many variations around it, all of them rely on Hessian vector multiplications. In our implementation we use the \textbf{minres} function from MATLAB to get the direction by using the Hessian-vector product. The pseudo code for Truncated Newton iteration to solve (\ref{eq:18}) is given in the Algorithm 3.

\subsection{Computational Complexity}
Assuming that kernel matrix $\bQ$ is stored in memory, then the computation of the loss term in (\ref{eq:10}) will require $O(l(d_{max}+ \log l))$ computation time. On the other hand the corresponding term in (\ref{eq:B}) require the computation time of $O(l^2)$. For large datasets it may not be feasible to store $\bQ$ in main memory. Therefore, for such datasets, Kernel rankSVM resorts to several block wise computations of $\bQ$, which may result in increased training time. This problem does not arise in our approach, as the maximum sub-matrix of $\bQ$ that it needs to store is of size $l \times d_{max}$.

\section{Empirical Evaluation}
In this section, we discuss the experimental evaluations of the proposed algorithm for sparse classifier design. In particular, we demonstrate that the proposed Sparse Kernel AUC algorithm results in a sparser classifier and gives comparable generalization performance with the Kernel rankSVM algorithm. Further, batch learning algorithms perform better than online learning algorithms on majority of real world datasets.
 
In our experiments, we used Gaussian kernel function, $K(x_{i},x_{j})$  = $exp(-\frac{1}{2\sigma^{2}}\| x_{i} - x_{j} \|^2)$ where $\sigma > 0$, for all the experiments. The kernel parameter $\sigma$ and regularization hyper-parameter C were tuned using cross-validation. For this, a grid of $(C,\sigma)$ values, where $C \in \{10^{-5}, 10^{-4}, ..., 10^{5}\}$ and $\sigma \in \{2^{-5}, 2^{-4}, ..., 2{^5}\}$ was searched. AUC performance corresponding to the $(C,\sigma)$ pair, which gave the best validation set AUC performance, is reported. The value $d_{max}$ was set to $\frac{l}{2}$. The proposed algorithm was terminated when $|J| = d_{max}$ was true or there was not significant change in the validation set AUC performance. All the experiments were performed using MATLAB implementations on a Intel(R) Xeon(R) CPU E5620@2.40GHz machine with 16 cores and 16 GB main memory under Linux. 

We compare the following methods: 1) Sparse Kernel AUC: our proposed sparse AUC optimization approach discuss in Section IV. 2) Kernel rankSVM: an extension of Kernel rankSVM method, discussed in \cite{kuo2014large}, to the AUC optimization problem. 3) Online Imbalanced Learning with Kernels (OILK)~\cite{yang2013online}, and 4) Adaptive Gradient Method for Online AUC Maximization (AdaOAM)~\cite{ding2016adaptive}. The performance of these methods was compared in terms of the AUC score on the test set (if a test set is available). If the test set is not explicitly available, AUC score on validation set, averaged over 4 independent run of five-fold splits of each dataset is reported. Since the aim of this paper is to design non-linear sparse classifier model using AUC optimization, we report the number of support vectors present in the final model for batch learning methods: Sparse Kernel AUC and Kernel rankSVM. The other two methods use an online learning approach and it may not be fair to compare the number of support vectors obtained using them with those obtained using batch learning methods. CPU time comparison of batch learning methods: Sparse Kernel AUC and Kernel rankSVM was not done as the implementations were done using MATLAB and C programming language respectively. But we compare computational complexity of both methods in Section IV-D.

We used 14 benchmark datasets to compare our proposed method, Sparse Kernel AUC, with other three methods. The dataset details are given in Table I. The datasets are available at \textbf{UCI}\footnote{\url{https://archive.ics.uci.edu/ml/datasets.html}} or \textbf{LIBSVM}\footnote{\url{https://www.csie.ntu.edu.tw/~cjlin/libsvmtools/datasets/}} dataset repository. Some multi-class datasets (glass, vechile, poker) were converted to class imbalanced binary datasets. As given in Table I, training+test splits are not available for some datasets.

\begin{table}[t]
\begin{center}
\scalebox{0.9}{
\begin{tabular}{|c|c|c|c|c|}

 \hline
 Datasets & $\# train \; inst$ &  $\# test\; inst$ & $\# dim$ & $n/p$ \\ 
 \hline \hline
 sonar & 208 & - & 60 & 1.144 \\
 \hline
 glass & 214 & - & 9 & 2.057 \\
 \hline
 ionosphere & 351 & - & 34 & 1.785 \\
 \hline
 balance & 625 & - & 4 &11.755 \\
 \hline
 australian & 690 & - & 14 & 1.247\\
 \hline
 vechile  & 846 & - & 18 & 3.251 \\ 
 \hline
 fourclass & 862 & - & 2 & 1.807 \\
 \hline
 svmguide3 & 1,243 & - & 22 &3.199 \\ 
  \hline
 a2a & 2,265 & - & 123 & 2.959\\ 
  \hline
 magic04 &19,020  & - &10 &1.843\\ 
  \hline
 segment & 210 & 2,100 & 19 & 6.000 \\
 \hline
 satimage & 4,435 & 2,000 & 36 & 9.279 \\
 \hline
 ijcnn1 & 49,990 & 91,701 & 22 & 10.0 \\ 
  \hline
 poker & 25,010 & 1,000,000 &11 &20.0 \\ 
  \hline
  
\end{tabular}}
\caption{Details of datasets}
\end{center}
\end{table}

\begin{table*}
\normalsize 
\begin{center}
\begin{tabular}{|c|c|c|c|c|} 
 \hline
 Datasets & Sparse Kernel AUC & Kernel rankSVM  & $OILK_{RS}$ & AdaOAM  \\ 
 \hline \hline
 sonar & 0.914 $\pm$ 0.043 \textbf{(105)} & \textit{0.951 $\pm$ 0.029} (167) & 0.929 $\pm$ .039 & -  \\ 
 \hline
 glass & 0.871 $\pm$ 0.054 \textbf{(150)} & \textit{0.881 $\pm$ 0.051} (171) & - & 0.816 $\pm$ 0.058  \\ 
 \hline
 ionosphere & 0.980 $\pm$ 0.017 \textbf{(182)} & \textit{0.987 $\pm$ 0.014} (281) & 0.954 $\pm$ 0.021 & -  \\ 
 \hline
 balance & \textit{1.000 $\pm$ 0.000} \textbf{(6)} & \textit{1.000 $\pm$ 0.000} (500) & - & 0.579 $\pm$ 0.106 \\ 
 \hline
 australian & 0.913 $\pm$ 0.034 \textbf{(256)} & \textit{0.930 $\pm$ 0.020} (552) & 0.925 $\pm$ 0.021 & 0.927 $\pm$ 0.016 \\ 
 \hline
 vechile  & 0.977 $\pm$ 0.022 \textbf{(431)} & \textit{0.995 $\pm$ 0.002} (677) & - & 0.818 $\pm$ 0.026 \\ 
 \hline
 fourclass  & 0.999 $\pm$ 0.000 \textbf{(108)} & \textit{1.000 $\pm$ 0.000} (690) & 0.829 $\pm$ 0.036 & - \\ 
 \hline
 svmguide3  & 0.823 $\pm$ 0.027 \textbf{(216)} & \textit{0.824 $\pm$ 0.026} (995) & - & 0.734 $\pm$ 0.038 \\ 
 \hline
 a2a  & \textit{0.880 $\pm$ 0.009} \textbf{(64)} & \textit{0.880 $\pm$ 0.010} (1741) & - & 0.873 $\pm$ 0.019  \\ 
 \hline
 magic04  & 0.874 $\pm$ 0.023 \textbf{(182)} & \textit{0.894 $\pm$ 0.007} (15124) & - & 0.798 $\pm$ 0.007  \\
 \hline
\end{tabular}
\end{center}
\caption{Validation set AUC Performance (mean $\pm$ std.) and maximum number of basis functions (in parenthesis) comparison of various methods. The AUC performance numbers for $OILK_{RS}$ and AdaOAM are reported from \cite{ding2016adaptive} and \cite{yang2013online} respectively.}
\end{table*}

\begin{table*}
\normalsize 
\begin{center}
\begin{tabular}{|c|c|c|c|c|} 
\hline
 Datasets & Sparse Kernel AUC & Kernel rankSVM & $OILK_{RS}$ & AdaOAM  \\
 \hline \hline
  segment & 0.996 \textbf{(91)} & \textit{0.998} (210) & 0.997 $\pm$ 0.003 & - \\
 \hline
 satimage & 0.961 \textbf{(431)} & \textit{0.969} (4435) & 0.896 $\pm$ 0.024& -\\
 \hline
 ijcnn1   & 0.995 \textbf{(182)} & \textit{1.000} (49990) & - & -  \\ 
 \hline
 poker & 0.668 \textbf{(363)} & \textit{0.674} (25010) & - & 0.571 $\pm$ 0.007 \\
 \hline
\end{tabular}
\end{center}
\caption{Test set AUC Performance and number of basis functions (in parenthesis) comparison of various methods. The AUC performance numbers for $OILK_{RS}$ and AdaOAM are reported from \cite{ding2016adaptive} and \cite{yang2013online} respectively.}
\end{table*}
\normalsize

\begin{figure*}
\begin{tabular}{c c c}
\includegraphics[scale = 0.32]{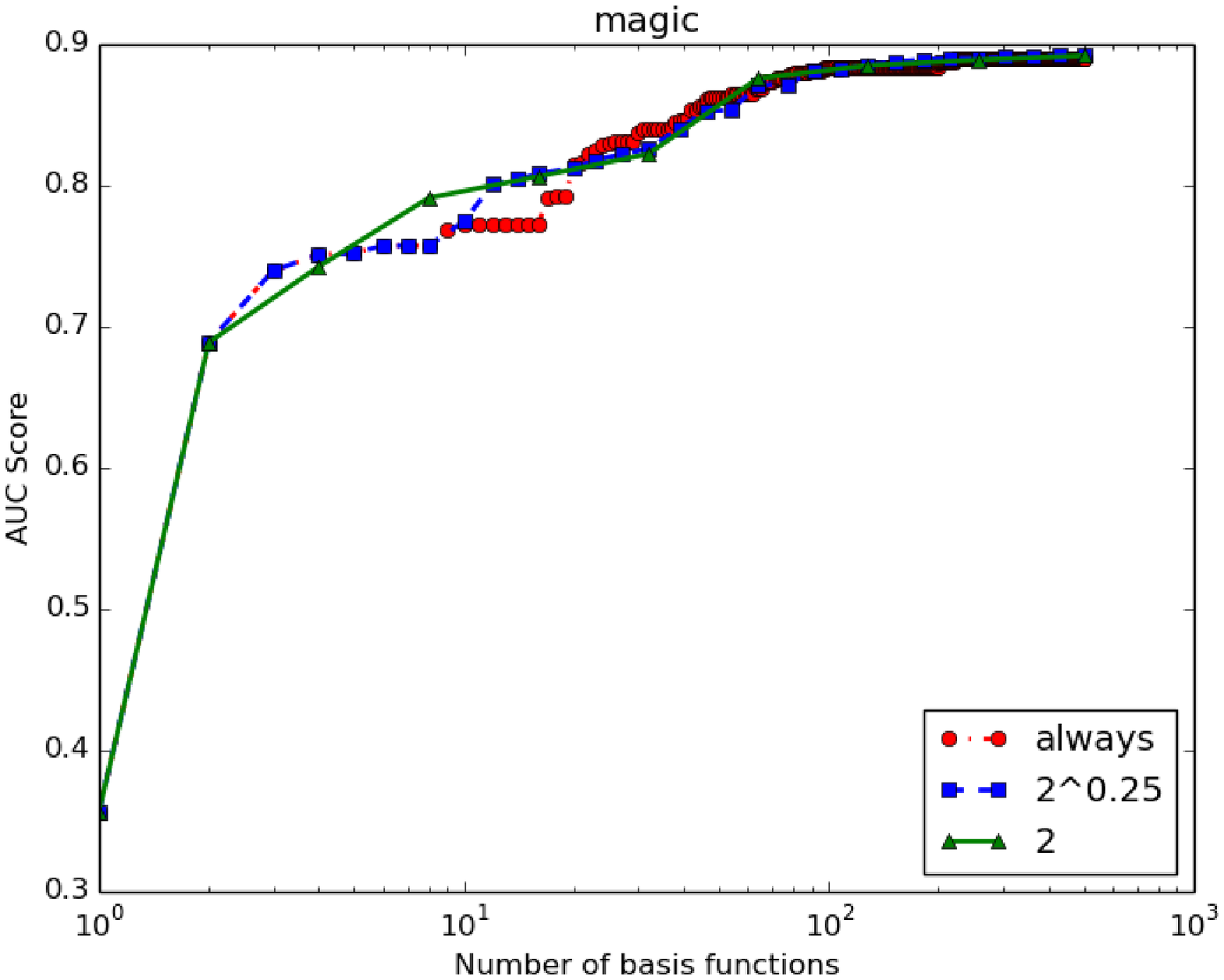} & 
\includegraphics[scale = 0.32]{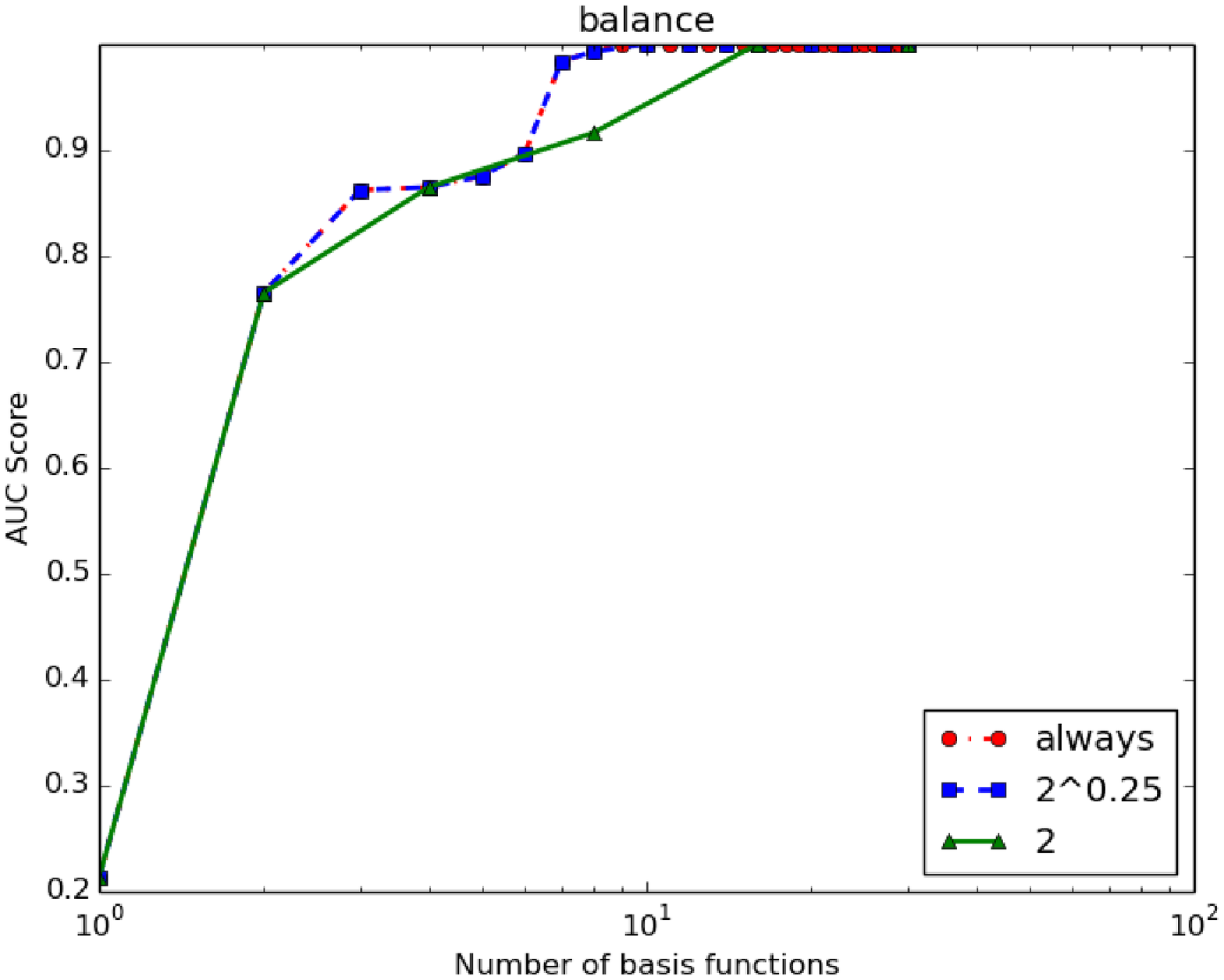} &
\includegraphics[scale = 0.32]{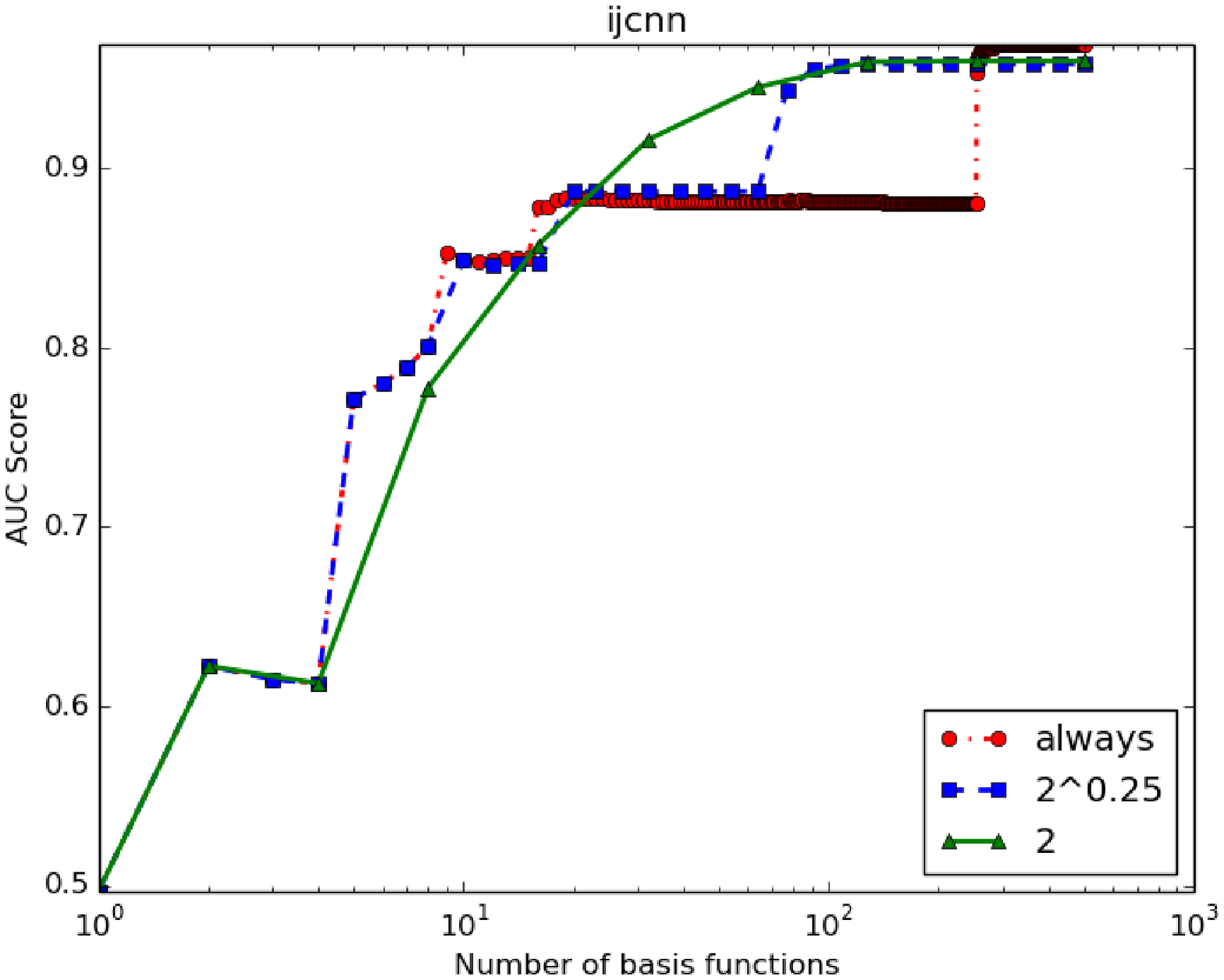} \\
\includegraphics[scale = 0.32]{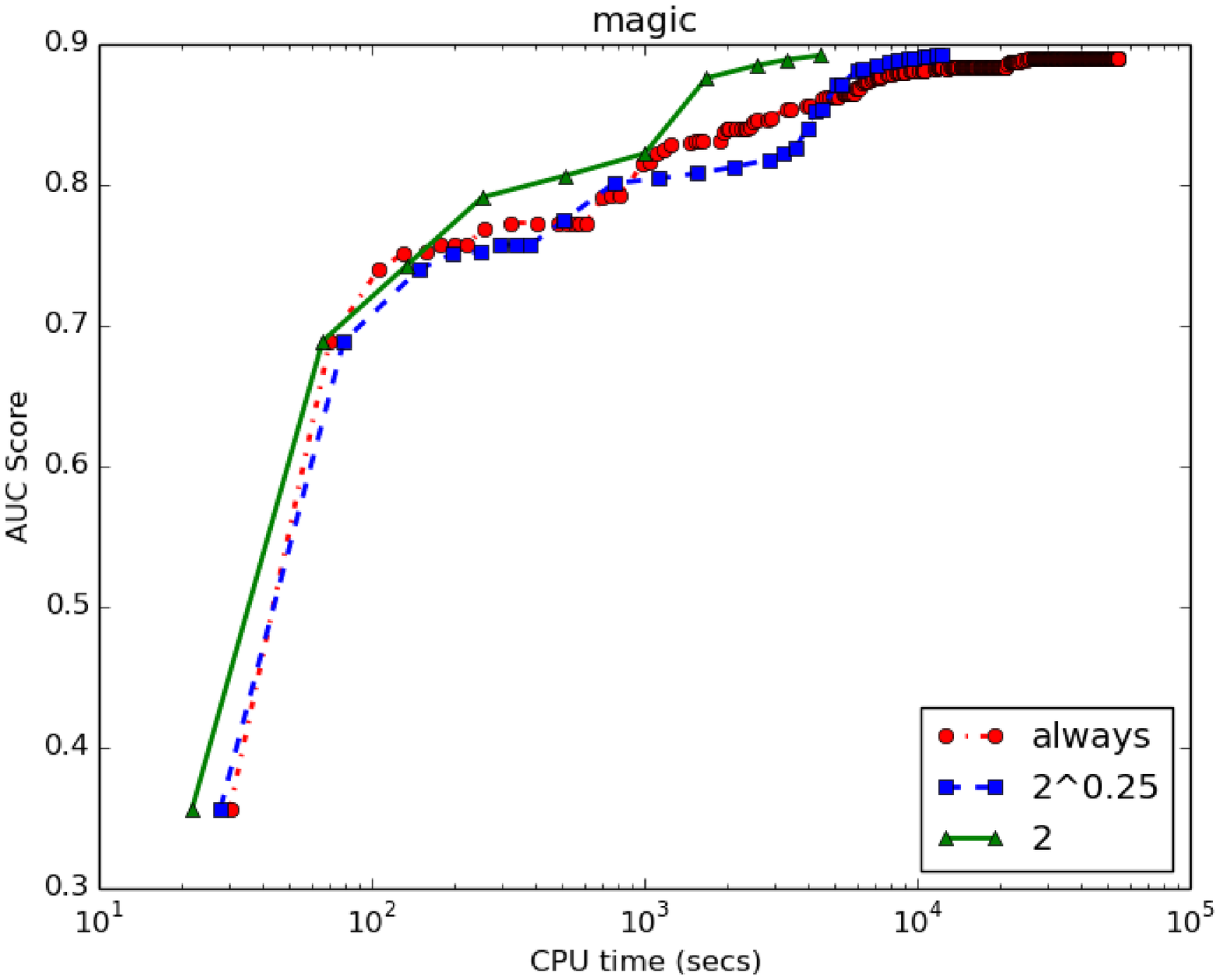}&
\includegraphics[scale = 0.32]{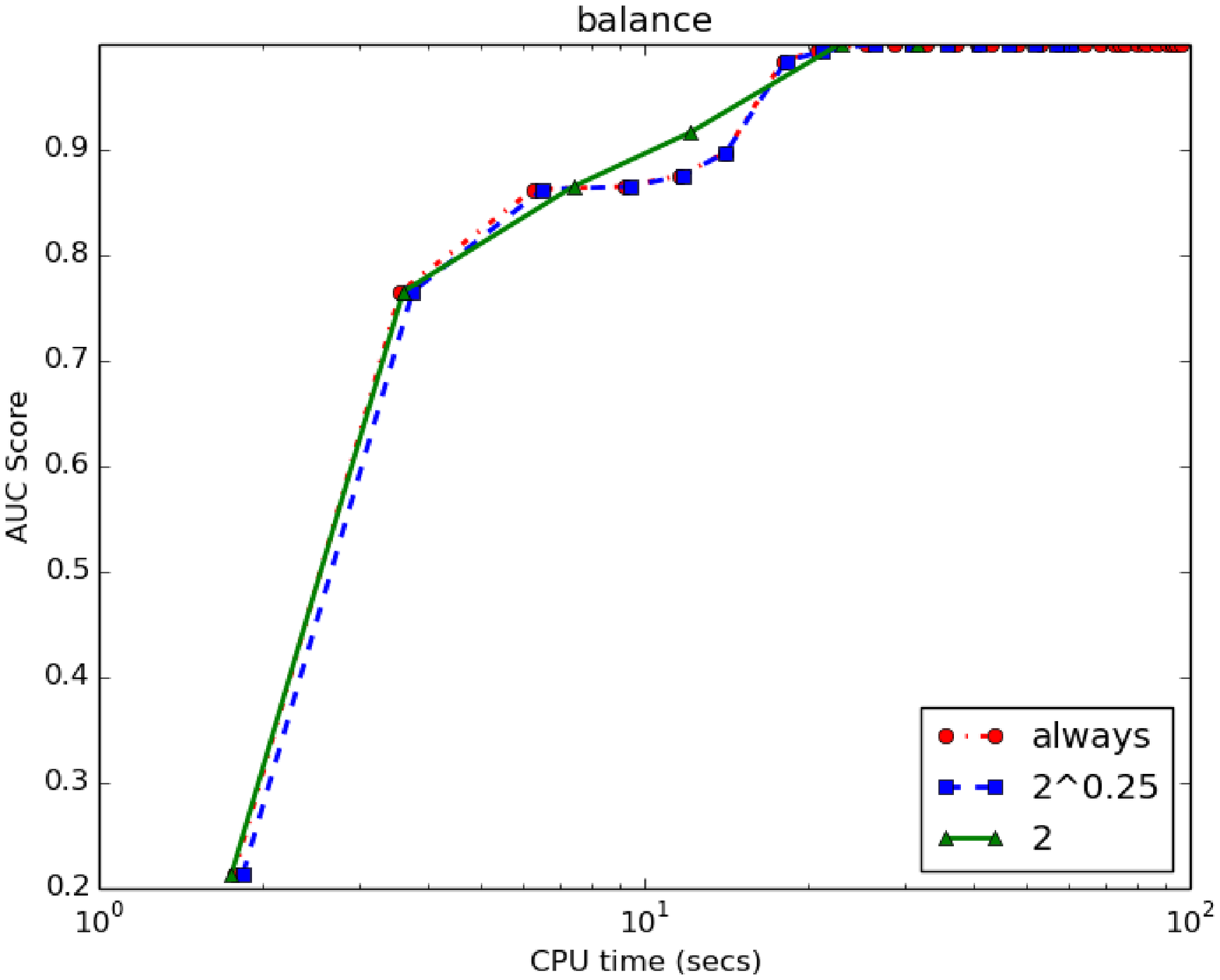}&
\includegraphics[scale = 0.32]{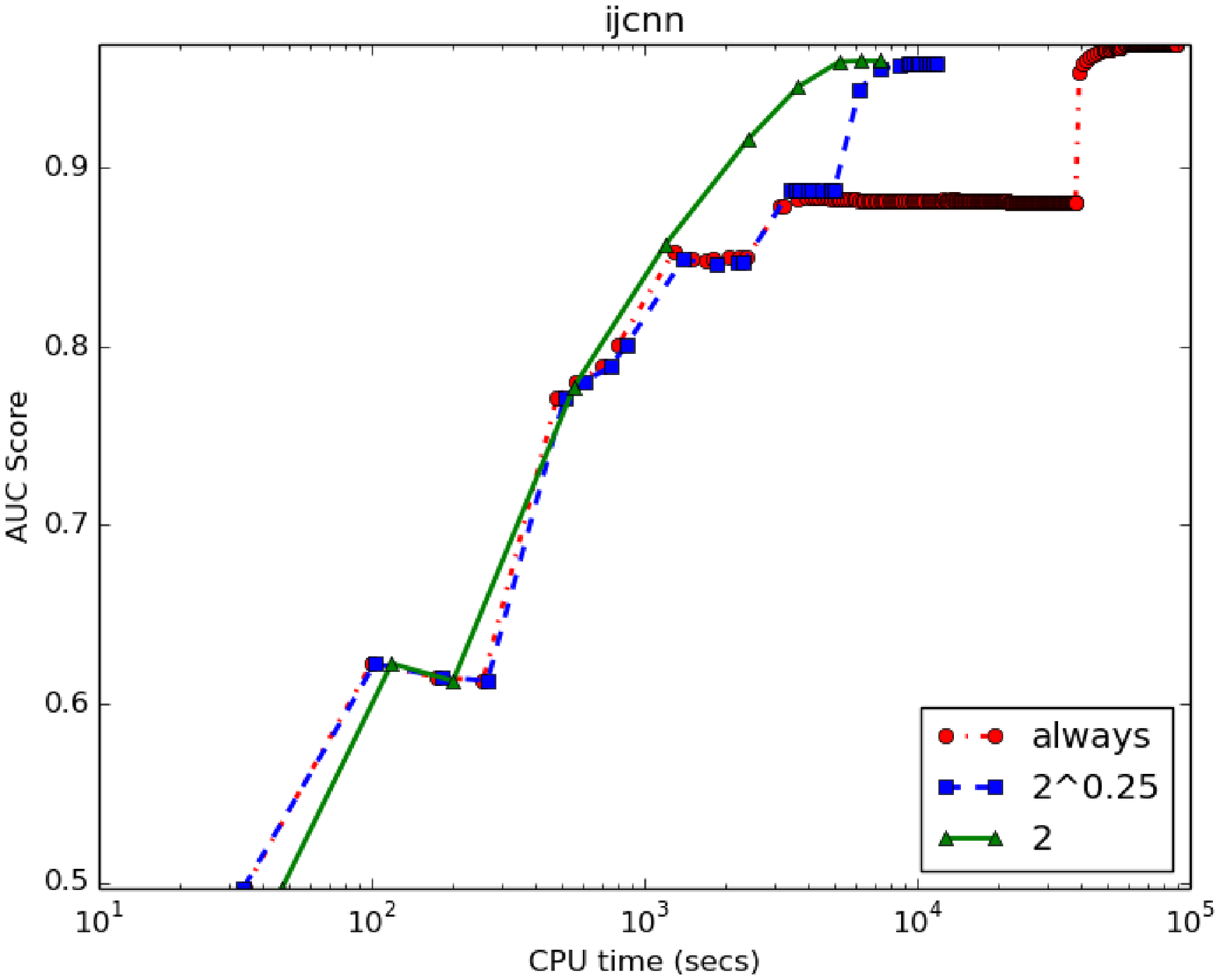}\\
\end{tabular}
\caption{Three different retraining strategies showing a different trade-off between AUC and time, always retraining is too time consuming.}
\label{fig:1}
\end{figure*}

{\bf Effect of retraining and $\kappa$} To make Algorithm 1 efficient, it may be a good idea to perform conjugate gradient optimization in step 5 only from time to time. We experimented with 3 retraining strategies where step 5 is executed after the addition of 1) every basis function (i.e always), 2) $|J| = \floor{2^{0.25}}$ basis functions and 3) $|J| = 2^{j},\;j=0,\ldots $, basis functions. The results are presented in figure (\ref{fig:1}). It is clear from this figure that, always retraining increases the training time. Similar generalization performance is achieved in other cases of retraining. We found that $\floor{2^{0.25}}$ was a good choice across many datasets and used it in our experiments.

\begin{figure*}
\begin{tabular}{c c c}
\includegraphics[scale = 0.32]{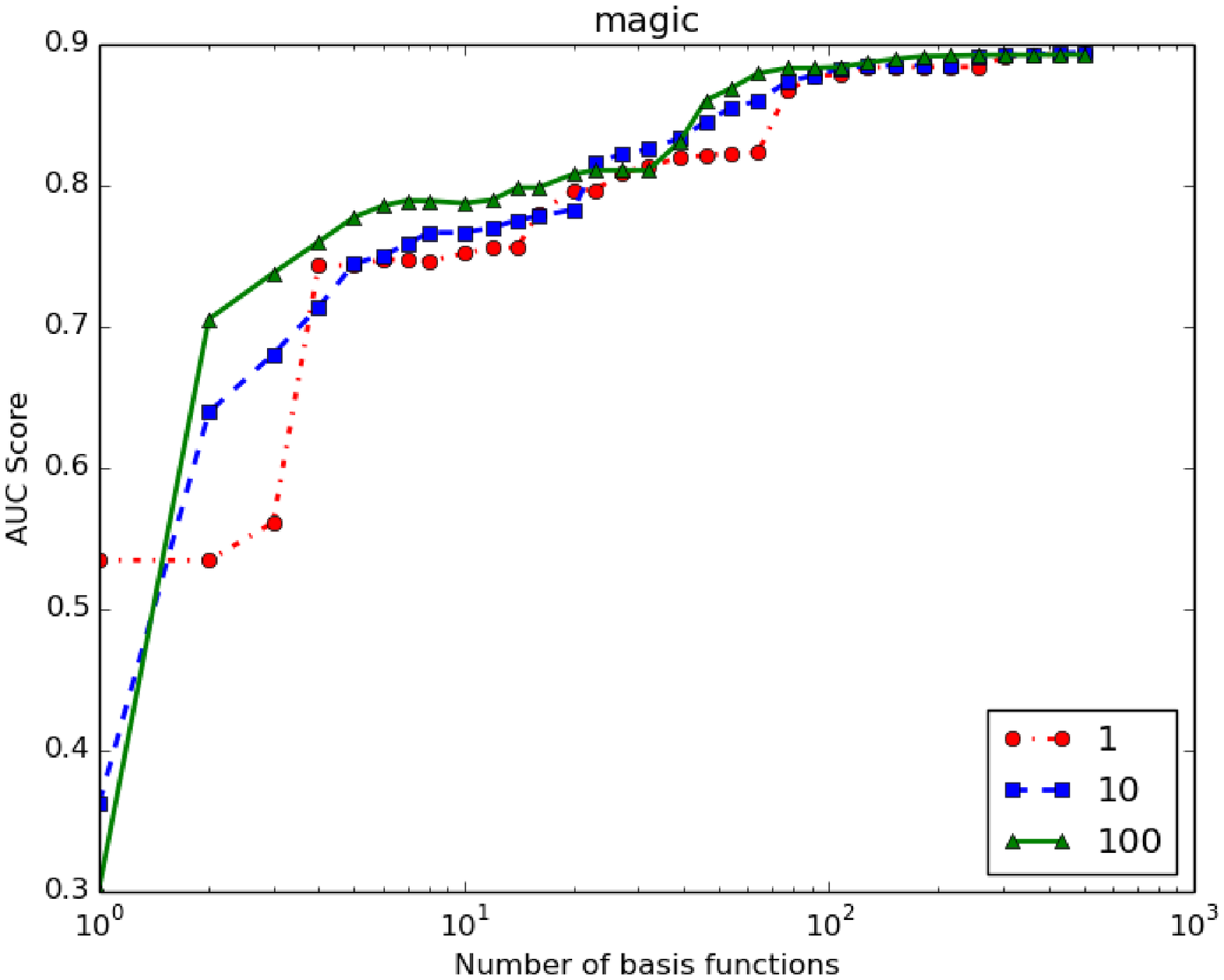} &
\includegraphics[scale = 0.32]{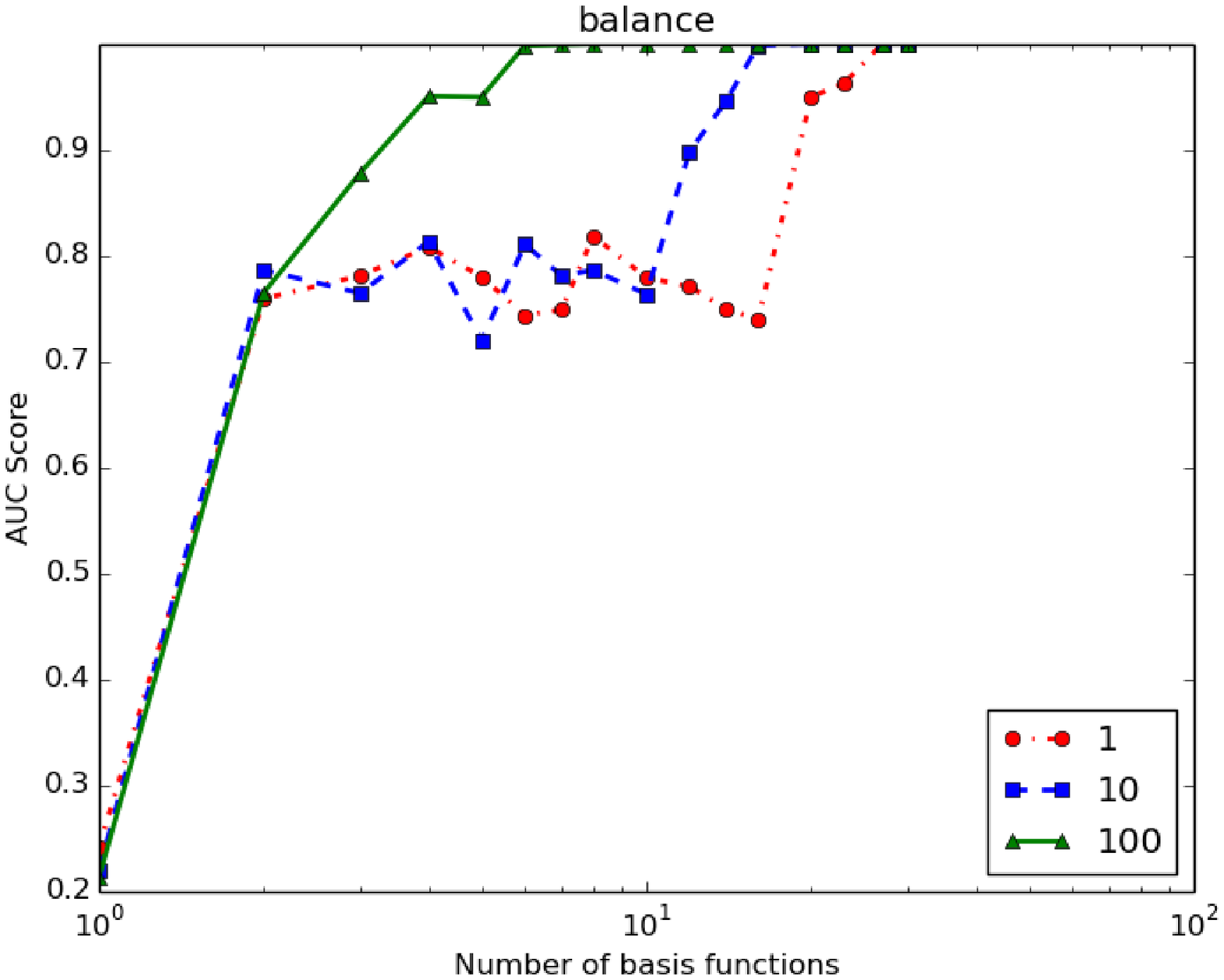} &
\includegraphics[scale = 0.32]{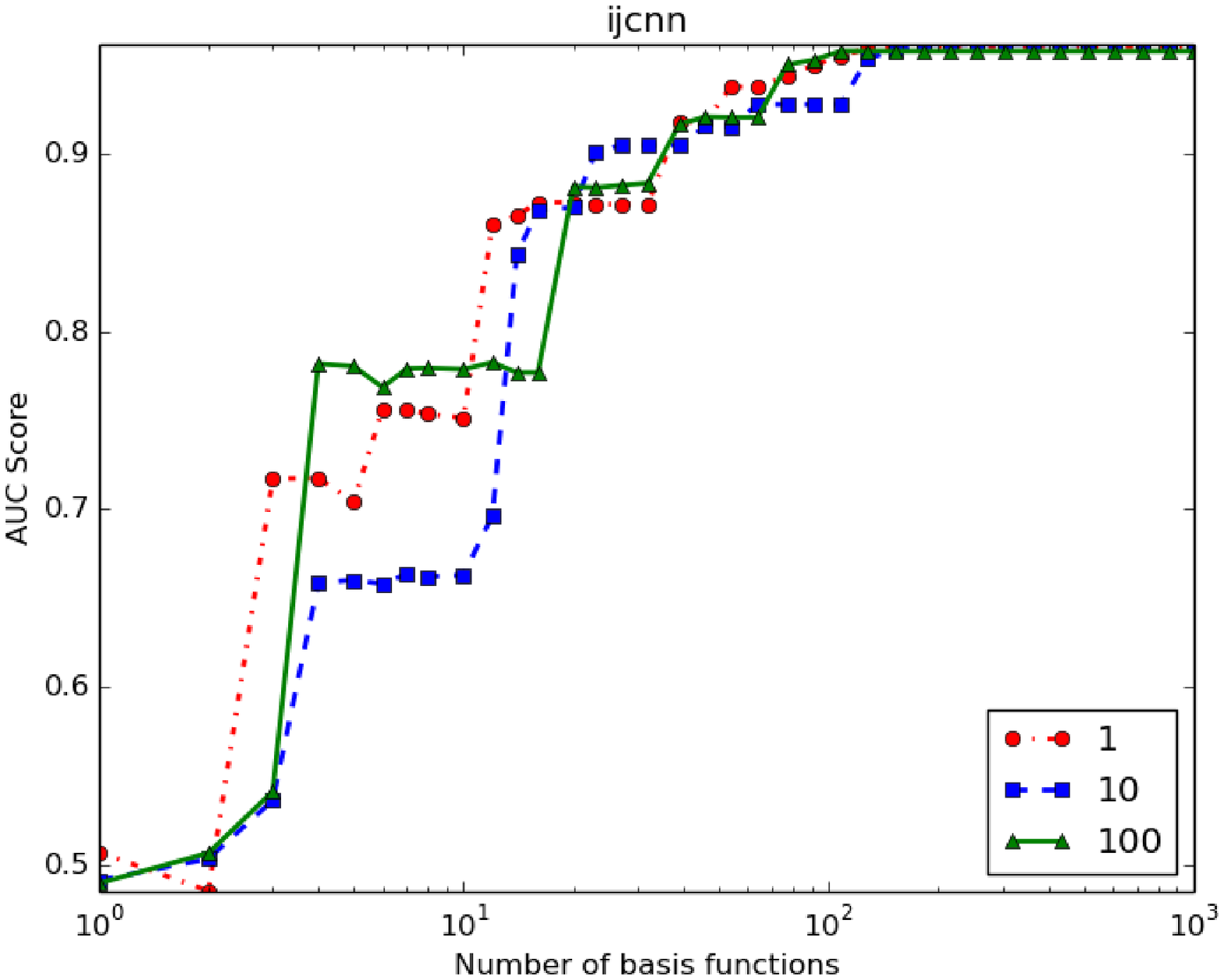} \\
\includegraphics[scale = 0.32]{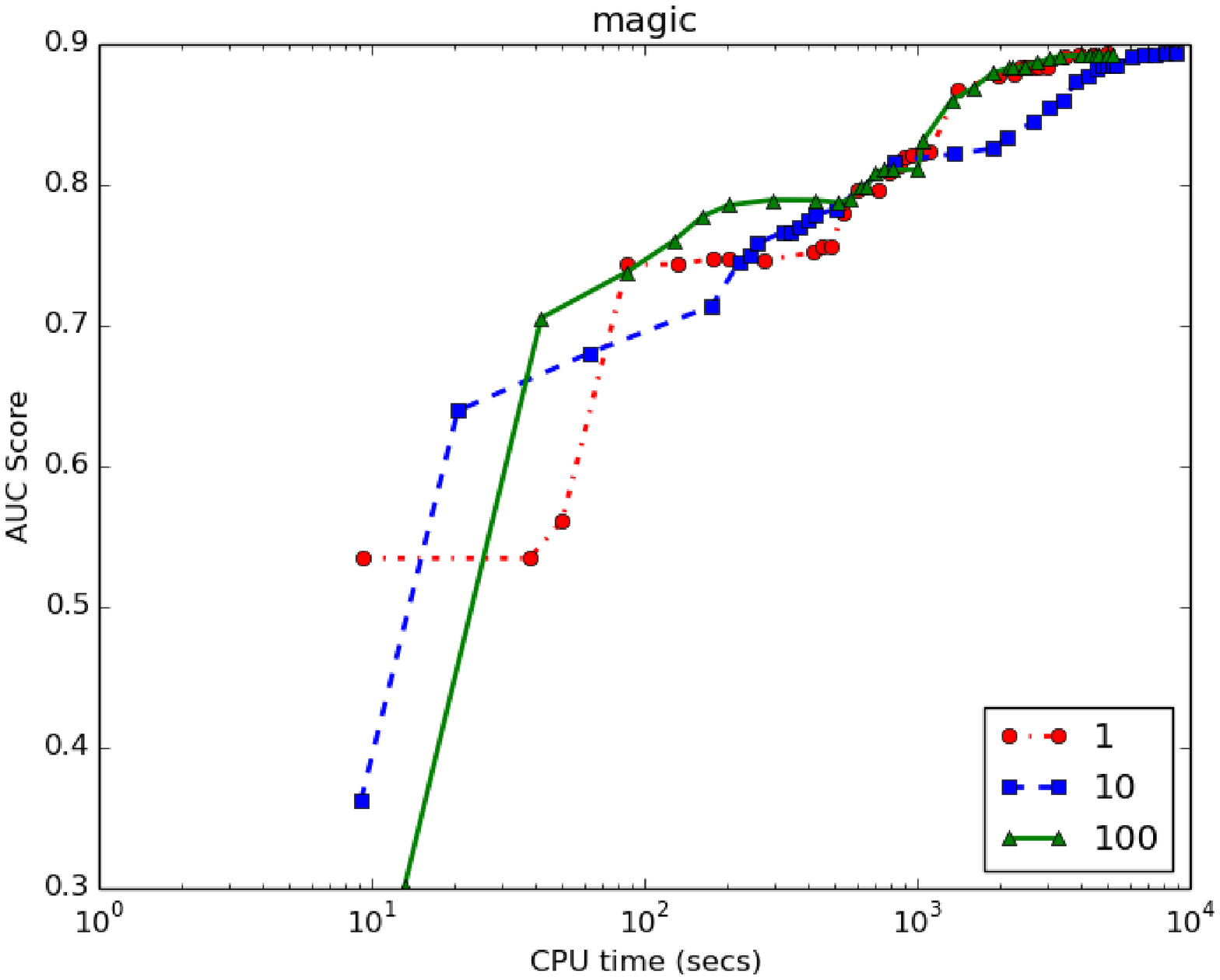} &
\includegraphics[scale = 0.32]{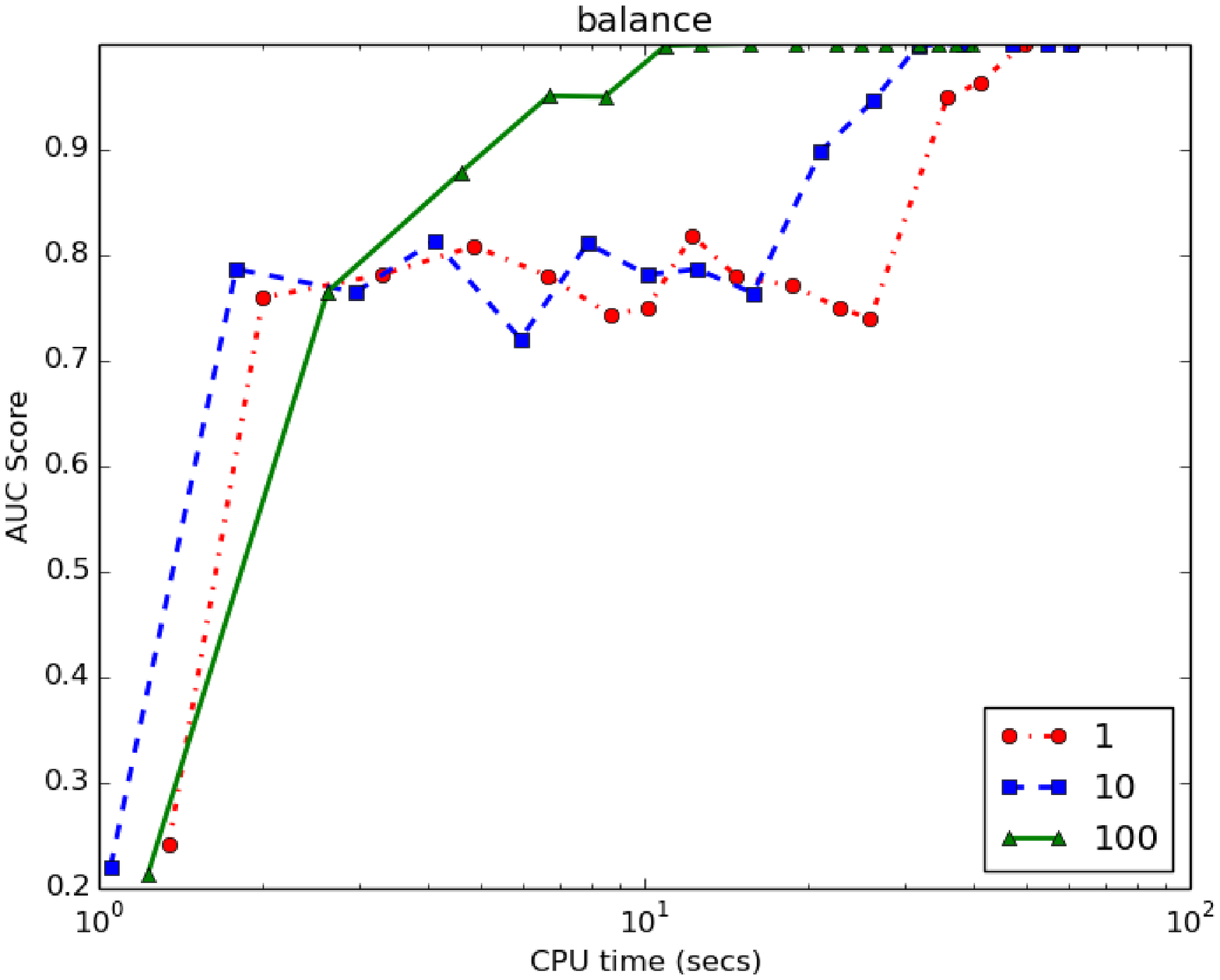} &
\includegraphics[scale = 0.32]{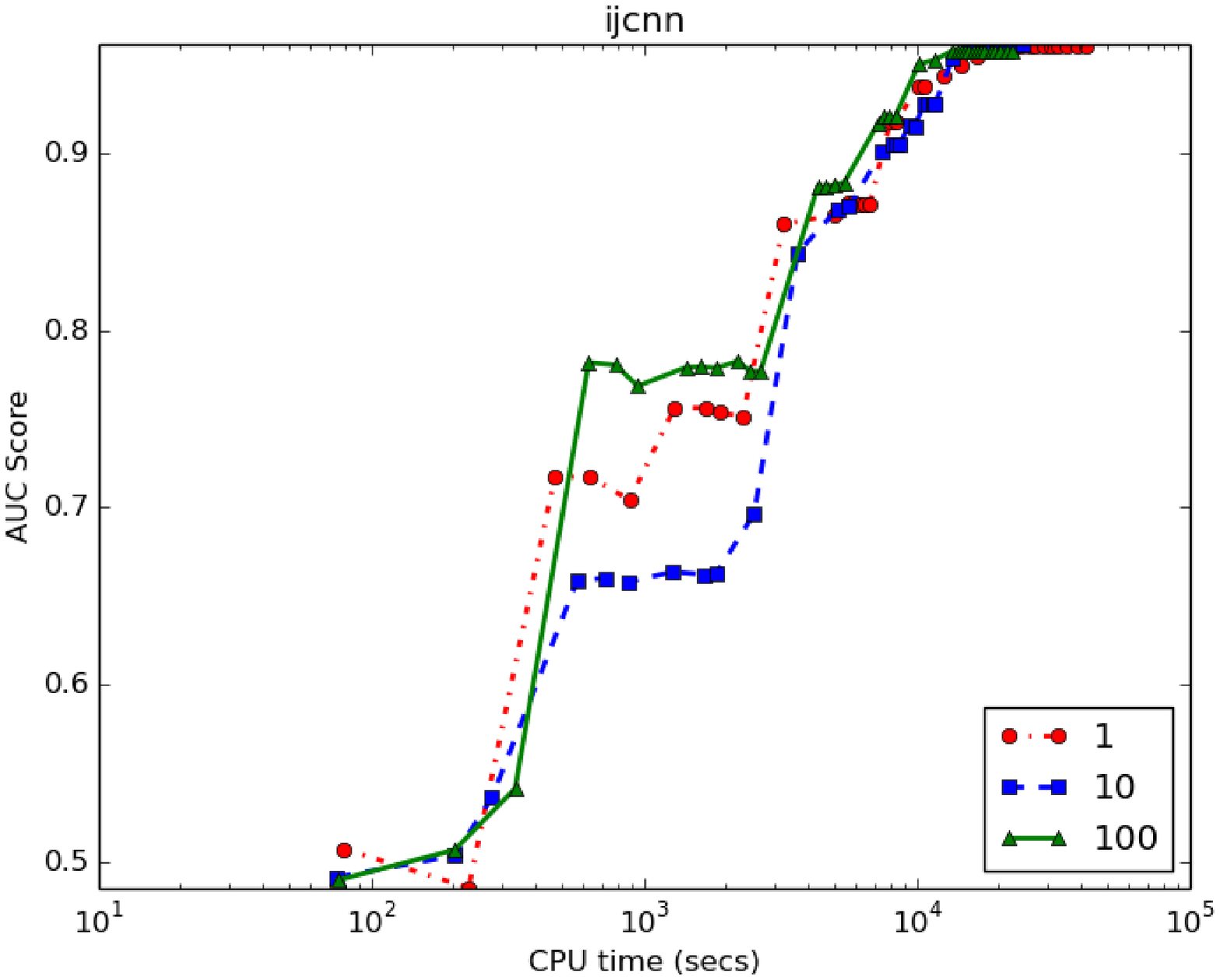}\\
\end{tabular}
\caption{Influence of the parameter $\kappa$: performance is not much affected, but the computational cost is a bit larger when $\kappa$=1. $\kappa$ = 100 seems a good trade-off.}
\label{fig:2}
\end{figure*}

As mentioned in section IV-B, instead of choosing a possible basis function from $\scd$ $\setminus J$, we chose a subset $\kappa$ of examples from this set as possible candidate for basis functions. Different values of $\kappa$ (1, 10, 100) were tried. The results are shown in figure (\ref{fig:2}). Although these 3 values of $\kappa$ resulted in similar steady state generalization performance, it was observed that for $\kappa=100$ steady state generalization performance was achieved faster. So, $\kappa=100$ was a good choice.

{\bf Discussion} From tables II and III, we observed that the generalization performance of the proposed Sparse Kernel AUC method is comparable with that of the Kernel rankSVM method. Both these batch learning methods perform significantly better than the OILK method on ionosphere, fourclass and satimage datasets.The kernel based methods, Sparse Kernel AUC, Kernel rankSVM and $OILK_{RS}$ perform better than linear classifier based method (AdaOAM) on majority of the datasets.

The proposed method required smaller number of basis functions than those required by Kernel rankSVM to achieve comparable AUC performance. Thus the proposed method is recommended for designing sparse classifiers for large datasets.

The reduction in the number of basis vectors is two orders of magnitude in case of some large datasets (magic, poker and ijcnn1).

\begin{figure}[H]
\includegraphics[scale = 0.45]{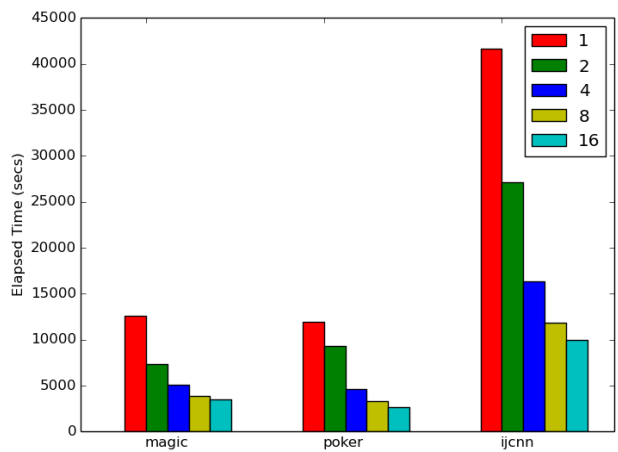}
\caption{Effect of increasing the numbers of core on time is shown for 3 benchmark datasets.}
\label{fig:3}
\end{figure}

{\bf Experiments in Multi-core Setting}
To study the speed-up of our proposed algorithm in multi-core environment, we parallelized steps 15-26 of Algorithm 2. The speed-up was studied on three large datasets by gradually increasing the number of cores from 1 to 16. Figure (\ref{fig:3}) depicts the time comparison. It is clear from this figure that significant speed-up can be obtained by running our method in multi-core environment. The speed-up is noticeably on large datasets like ijcnn1. Detailed investigation is however needed to study the parallelization of the complete proposed algorithm.

\section{Conclusion} 
This paper studied a new and efficient learning algorithm to design a sparse nonlinear classifier using AUC maximization. The algorithm tackles the challenge of lengthy training times of kernel methods by greedily adding the required number of basis functions in the model. We demonstrated that the resulting sparse classifier achieved comparable generalization performance with that achieved by full models. On many large datasets, it was observed that the proposed algorithm results in using significantly small number of basis functions in the model. We also demonstrated that batch learning algorithms for AUC optimization perform better than online algorithms on many datasets. We are currently investigating the extension of these ideas to a distributed setting. The MATLAB code for this paper is available at this dropbox link  \url{https://www.dropbox.com/s/ha7w3o029lhb5bn/ICDM_AUCCode.tar.gz?dl=0 }

\bibliographystyle{plain}
\bibliography{refer}

\end{document}